\pdfoutput=1
\documentclass[11pt]{article}

\usepackage[]{acl}

\usepackage{times}
\usepackage{latexsym}
\usepackage{graphicx}
\usepackage{subcaption}
\usepackage{caption}
\usepackage{xcolor}
\usepackage{multicol}
\usepackage{multirow}
\usepackage{tabularx}
\usepackage{amsmath}
\usepackage{amssymb}
\usepackage{booktabs}
\usepackage{float}
\usepackage{mathtools}
\usepackage{cleveref}
\usepackage{adjustbox}
\usepackage[T1]{fontenc}
\usepackage[utf8]{inputenc}

\usepackage{microtype}

\title{DialoKG: Knowledge-Structure Aware\\ Task-Oriented Dialogue Generation}
 \author{Md Rashad Al Hasan Rony$^{1,3}$, \textbf{Ricardo Usbeck}$^{2}$\textbf{, Jens Lehmann}$^{1,3}$ \\ {$^{1}$University of Bonn, $^{2}$University of Hamburg, $^{3}$Fraunhofer IAIS Dresden} \\ \texttt{\{rashad.rony,jens.lehmann\}@iais.fraunhofer.de}\\\texttt{\{lehmann\}@uni-bonn.de}\\\texttt{ricardo.usbeck@uni-hamburg.de}}

\begin{document}
\maketitle
\begin{abstract}
%Knowledge-grounded dialogue systems have recently gained substantial research attention. 
%In a task-oriented setting, such systems focus on achieving a specific goal.
Task-oriented dialogue generation is challenging since the underlying knowledge is often dynamic and effectively incorporating knowledge into the learning process is hard. It is particularly challenging to generate both human-like and informative responses in this setting. Recent research primarily focused on various knowledge distillation methods where the underlying relationship between the facts in a knowledge base is not effectively captured. In this paper, we go one step further and demonstrate how the structural information of a knowledge graph can improve the system's inference capabilities. Specifically, we propose DialoKG, a novel task-oriented dialogue system that effectively incorporates knowledge into a language model. Our proposed system views relational knowledge as a knowledge graph and introduces (1) a structure-aware knowledge embedding technique, and (2) a knowledge graph-weighted attention masking strategy to facilitate the system selecting relevant information during the dialogue generation. An empirical evaluation demonstrates the effectiveness of DialoKG over state-of-the-art methods on several standard benchmark datasets.
\end{abstract}

\section{Introduction}

%\about task oritented dialogues:
%Generating informative dialogues have recently gained attention because its wide range of application.
%Knowledge-grounded dialogue systems aim to generate informative dialogues in a multi-turn settings, based on the available domain knowledge.
\begin{figure}[!ht]
\centering
    \includegraphics[width=\columnwidth]{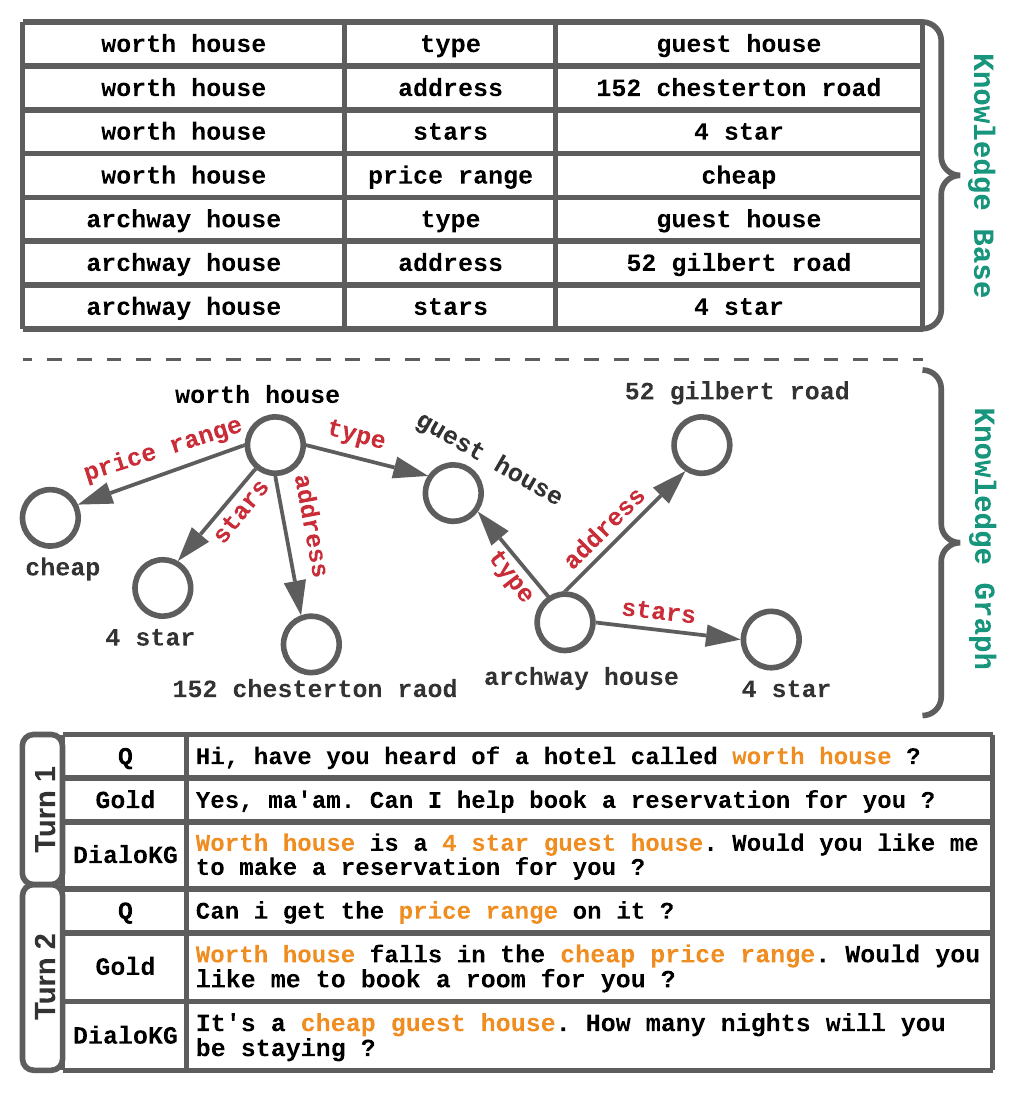}
    \caption{An illustration of knowledge-based multi-turn dialogue where DialoKG models the knowledge base as a Knowledge Graph. % Here, \textbf{Q} refers to the user question and \textbf{Gold} refers the ground truth response and the words in orange color are from the knowledge graph.
    The user utterance is denoted by \textbf{Q}, the ground-truth response by \textbf{Gold}, and the words in \textcolor{orange}{orange} are knowledge graph entries.} 
    \label{fig:intro}
\end{figure}
Traditional task-oriented dialogue systems are designed to achieve specific goals such a restaurant reservation, hotel booking and car navigation. These systems are often empowered by external domain- or task-specific knowledge that enables them to generate informative dialogues~\cite{eric-etal-2017-key,wu2019global,qin-etal-2019-entity,9053667}. The external knowledge in these systems is usually incorporated in the form of structured knowledge triples~\cite{zhou2018commonsense,liu2018knowledge} or unstructured documents~\cite{ye2020knowledge,Ghazvininejad}. Figure~\ref{fig:intro} depicts a knowledge-grounded dialogue about reserving a hotel. %The majority of recent works concentrate on selecting the relevant knowledge from the provided knowledge base and generating a response based on the given context and selected knowledge

Recent research primarily concentrated on various knowledge filtering methods for selecting relevant knowledge~\cite{wen-etal-2018-sequence,Kim2020ICLR,wu-etal-2020-improving-knowledge}.
%These approaches mainly leverage Memory Pointer Network~\cite{memnet,mem2seq,wu2019global}, Copy Network~\cite{gu-etal-2016-incorporating,lin-etal-2020-generating,chaudhuri2019using} and similarity based knowledge distillation techniques~\cite{wen-etal-2018-sequence,raghu-etal-2021-constraint} for knowledge selection and dialogue generation. 
\begin{figure*}[ht]

\centering
  \begin{subfigure}[b]{0.56\textwidth}
    \includegraphics[width=0.9\textwidth]{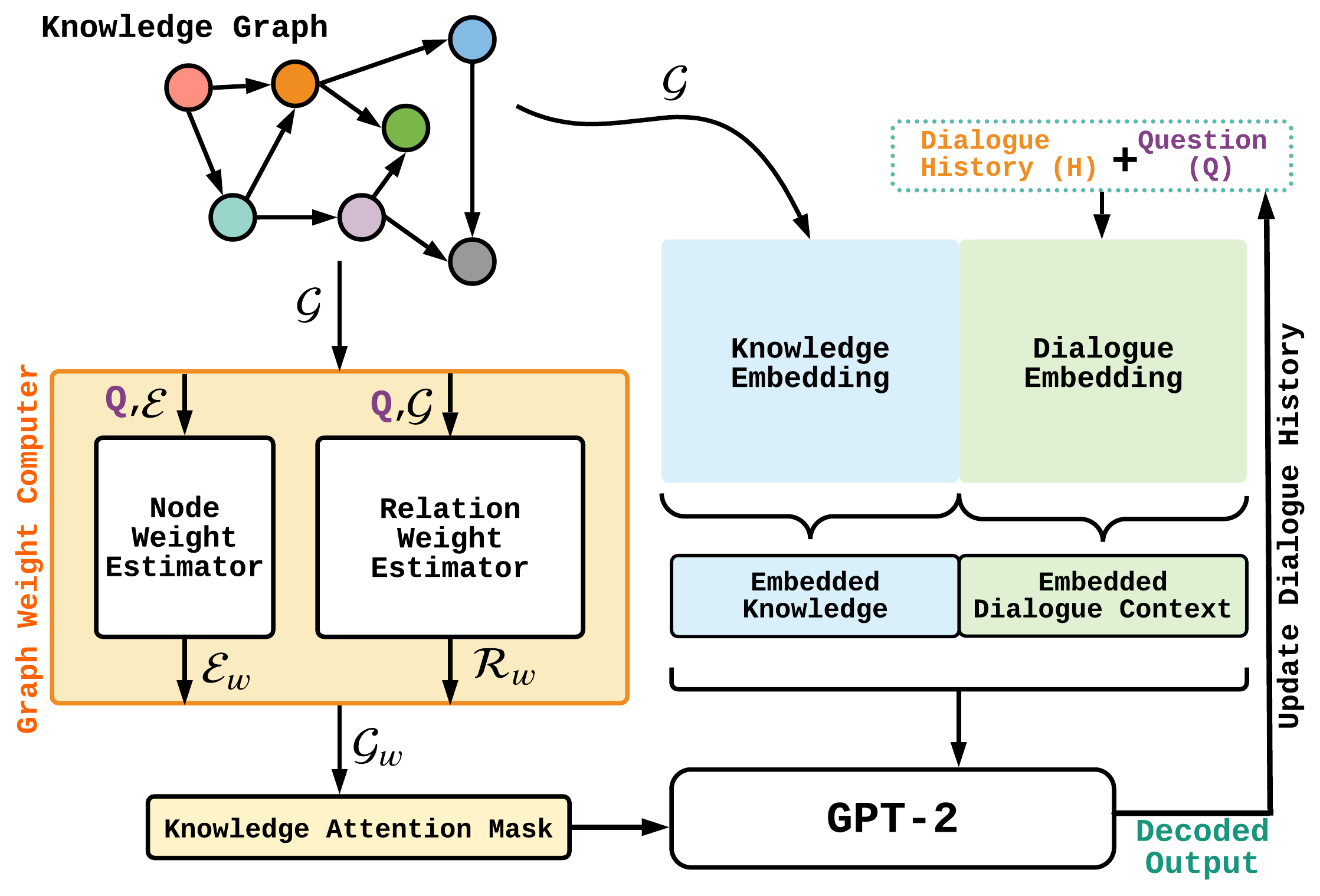}
    \caption{System architecture.}
    \label{fig:arch}
  \end{subfigure}
  \hfill
  \begin{subfigure}[b]{0.4\textwidth}
    \includegraphics[width=0.9\textwidth]{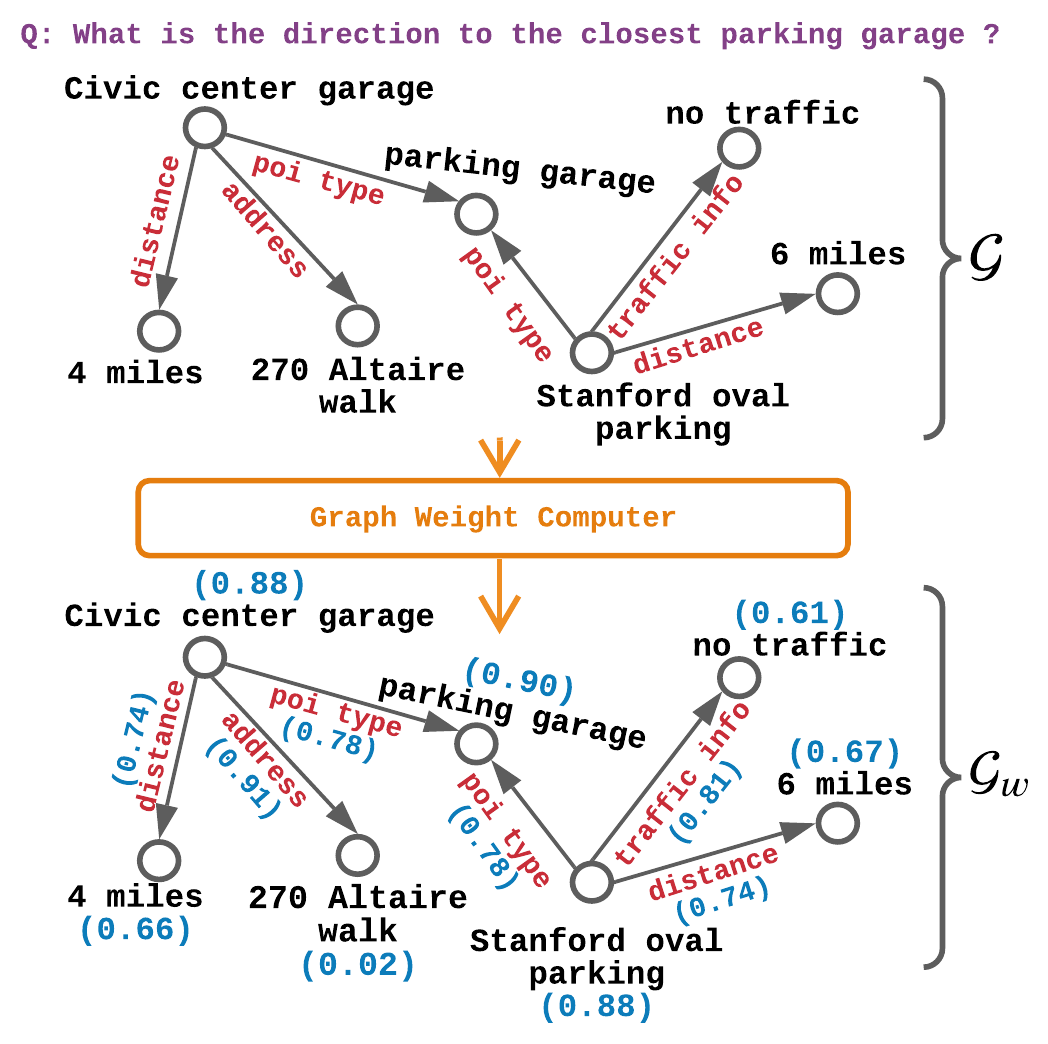}
    \caption{Weighted-graph computation.}
    \label{fig:graphwg}
  \end{subfigure}
  \caption{A high-level overview of DialoKG is shown in Figure (a). Figure (b) depicts the input and output of the \textit{Graph Weight Computer} module of DialoKG.}

\end{figure*}
%However, the similarity based knowledge selection approaches do not effectively handles the relationship between the knowledge entries and thus often lead to incorrect knowledge selection and dialogue generation. 
These approaches treat the knowledge triples independently and leverage Pointer Networks and copy mechanisms to generate knowledge-grounded dialogues~\cite{pointernet,gu-etal-2016-incorporating,wu2019global,memnet,raghu-etal-2021-constraint,chaudhuri2021grounding}% and focus on capturing the structural information of the dialogue
. Typically, these systems generate a template or sketch-response during training and learn to fill in the slots with knowledge graph entries. % The underlying semantics of a knowledge graph, such as the relationship between entity and relation, is not captured in these approaches. Because of this limitation, they often generate incorrect and noisy dialogues~\cite{madotto-etal-2020-learning}. 
Such systems face two issues when they try to generate dialogues in a multi-domain setting. \textbf{Firstly}, they are unable to capture the underlying semantics of a knowledge graph, such as the relationship between entity and relation. This leads frequently to incorrect and inappropriate dialogue generation~\cite{lin-etal-2020-generating}. \textbf{Secondly}, they lack the ability to encode dynamic knowledge in a multi-domain setting, resulting in noisy dialogues~\cite{madotto-etal-2020-learning}. %Generally, generating knowledge-grounded and coherent dialogue at the same time is a challenging task.
Generally, integrating a knowledge base into the learning process and generating correct and coherent dialogues at the same time is a challenging task.
%However, similarity-based knowledge selection approaches do not effectively handle the relationship between the knowledge entries, resulting in inaccurate dialogue generation.  However, these approaches primarily focus on capturing the structural information of the dialogue and context, rather than the provided knowledge.

%Addressing the issues mentioned above, 
In this paper, we propose a novel task-oriented dialogue system, named DialoKG that employs structural information of the knowledge graph into a language model (LM) for generating informative dialogues (see Figure~\ref{fig:arch}). 
%For the dialogue generation task, w
For this purpose, we exploit GPT-2~\cite{radford2019language} - a language model developed based on a stack of Transformer decoders~\cite{vaswani2017attention}.
%We exploit GPT-2~\cite{radford2019language}, a language model based on a stack of transformer decoders~\cite{vaswani2017attention}, for dialogue generation. 
%DialoKG utilize GPT-2 as a base for generating human-like and engaging dialogues. 
Specifically, we introduce a novel structure-aware multiple embedding layer-based knowledge embedding technique that constructively embeds the underlying relationship between the knowledge triples %(see Figure~\ref{fig:enc})
.
%, based on the properties of the knowledge entries. 
DialoKG interprets the knowledge as a knowledge graph% (see Figure~\ref{fig:intro})
; therefore, separate embedding layers for word token, entity, triple and token type enable the system to capture the graph features (e.g., subject, relation and object). This enables the system to generate correct and human-like dialogues and prevents generating erroneous responses such as "\textit{4 miles is located at 792 Bedoin Street Starbucks away}". Furthermore, the ability to correctly capture the relationship in the knowledge graph eliminates the need for template-based or sketch-based response generation. %Template-based systems are trained to fill in the slots in the response template with the words from knowledge entries without learning implicit relationship of the knowledge entries.

In order to guide the decoder on relevant parts of the knowledge graph, we propose a new knowledge attention masking method.
%A knowledge attention masking method is also proposed based on a weighted-graph, to guide the decoder in selecting relevant knowledge during the dialogue generation process.
For constructing the knowledge attention mask, in each dialogue turn, a weighted graph is computed in two steps: 1) Entity weights are computed using a pre-trained language model that estimates the importance of an entity for the given utterance, and 2) relation weights are computed based on the concept of graph convolution networks (GCN)~\cite{kipf2017semi}. Both steps take the user utterance into consideration, i.e., the obtained weighted graph is question specific. % (details in \textsection{\ref{subsec:relweight}}).
A set of triples is then selected based on the most relevant entities and relations of the weighted graph to construct a knowledge attention mask for the language model. This allows the masked language model to focus on relevant graph triples. We hypothesise that this leads to the generation of more accurate responses and enhance the model's capabilities of understanding the domain and task.%, given the current context and the user question.

To assess the performance of DialoKG, we conduct experiments on three public benchmarks: SMD~\cite{eric-etal-2017-key}, CamRest~\cite{wen-etal-2017-network} and Multi-WOZ 2.1~\cite{budzianowski-etal-2018-multiwoz}. We evaluate the system generated responses using both human and automatic metrics. Furthermore, we analyse impact of the individual components on the overall performance to verify the effectiveness.
%The results demonstrate that DialoKG brings several advantages for generating accurate, knowledge-grounded and engaging dialogues (see \textsection{\ref{sec:results}}). 
%DialoKG achieves 3\%, 4\% and 3.9\% improvement on SMD, Camrest and Multi-WOZ 2.1 datasets Entity-F1 
Our experimental results show that DialoKG outperforms state-of-the-art models in knowledge-grounded dialogue generation and can generate human-like responses. We made our code publicly available~\footnote{\url{https://github.com/rashad101/DialoKG}}.

\begin{figure*}[ht]
    \centering
    \includegraphics[width=\textwidth]{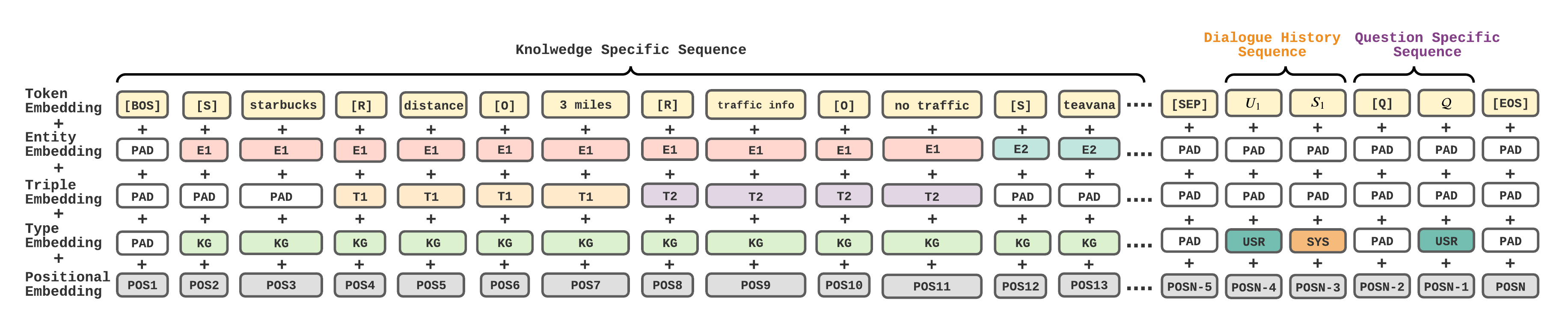}
    \caption{An illustration of knowledge and dialogue embedding techniques.}
    \label{fig:enc}
\end{figure*}

\section{Approach}
\subsection{Problem Definition}
DialoKG aims to generate informative responses given a dialogue history, a question and a knowledge base. We define the dialogue history $\mathcal{H}$ as a set of turns between two speakers, such that $\mathcal{H}=\{U_{1},S_{1},..,U_{t},S_{t}\}$, where $U_{i}$ and $S_{i}$ are the sequences of words in turn $i$. We assume that the knowledge is stored in a multi-relational knowledge graph $\mathcal{G}$. Here, $\mathcal{G}$ is a set of triples $\mathcal{T}$ such that $\mathcal{T}$ $\subseteq$
$\mathcal{E} \times \mathcal{R} \times \mathcal{E}$, where 
$\mathcal{E}$ is the set of entities and $\mathcal{R}$ the set of relations.
A triple $\mathcal{T} \in \mathcal{G}$ is denoted as ($s$, $r$, $o$) in which $s \in\mathcal{E}$ and $o \in\mathcal{E}$ denote the subject and object entities,
respectively, and $r \in \mathcal{R}$ is the relation between them. 
%, where $s$, $o\in\mathcal{E}$ and $r\in \mathcal{R}$. 
We use the terms "Knowledge Graph" and "Graph" interchangeably throughout this paper. Furthermore, we denote the user utterance of the current dialogue turn as $\mathcal{Q}$.  A GPT-2~\cite{radford2019language} language model is used in this paper to generate responses. However, any Transformer decoder-based LM can be used. Formally, the probability distribution of generating a response by the language model is defined as:
\begin{equation}
\small
    p(S_{t}|\mathcal{H},\mathcal{Q}, \mathcal{G}) = \prod_{i=1}^{n}  p(s_{i}|s_{1},.,s_{i-1},\mathcal{H},\mathcal{Q},\mathcal{G})
\end{equation}
Here, $S_{t}$ is the generated response in turn $t$ %, where $S_{t}$=$\{s_{1},..,s_{n}\}$ 
and $n$ is the maximum length of the generated response.
%We define the dialogue context $\mathcal{C}$, where $\mathcal{C}$=[$\mathcal{H};\mathcal{Q}$].

\subsection{Knowledge and Dialogue Embedding}
\label{subsec:inpemb}
DialoKG % follows the encoder-decoder design paradigm. The encoder takes a
takes a knowledge graph $\mathcal{G}$, dialogue history $\mathcal{H}$, and the current user utterance $\mathcal{Q}$ together as input and constructs a single input sequence as depicted in Figure~\ref{fig:enc}. The first part of the sequence %(\textit{Knowledge Encoding})
contains graph related information (i.e., subject, relation, and object) and the latter part dialogue specific information such as dialogue history ($\mathcal{H}$) and the current user utterance ($\mathcal{Q}$).%on encoding the dialogue related information.
\begin{figure*}[ht]

\centering
  \begin{subfigure}[b]{0.2\textwidth}
    \includegraphics[width=\textwidth]{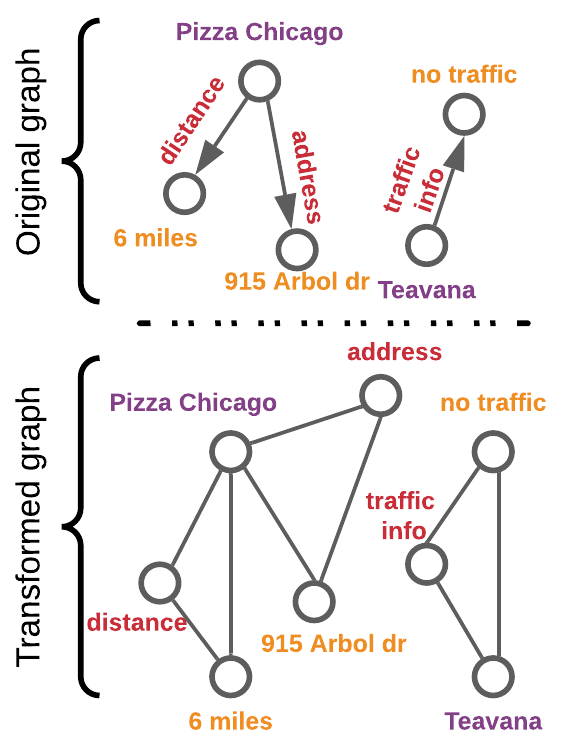}
    \caption{Graph transformation.}
    \label{fig:lapgraphmini}
  \end{subfigure}
  \hfill
    \begin{subfigure}[b]{0.79\textwidth}
    \includegraphics[width=\textwidth]{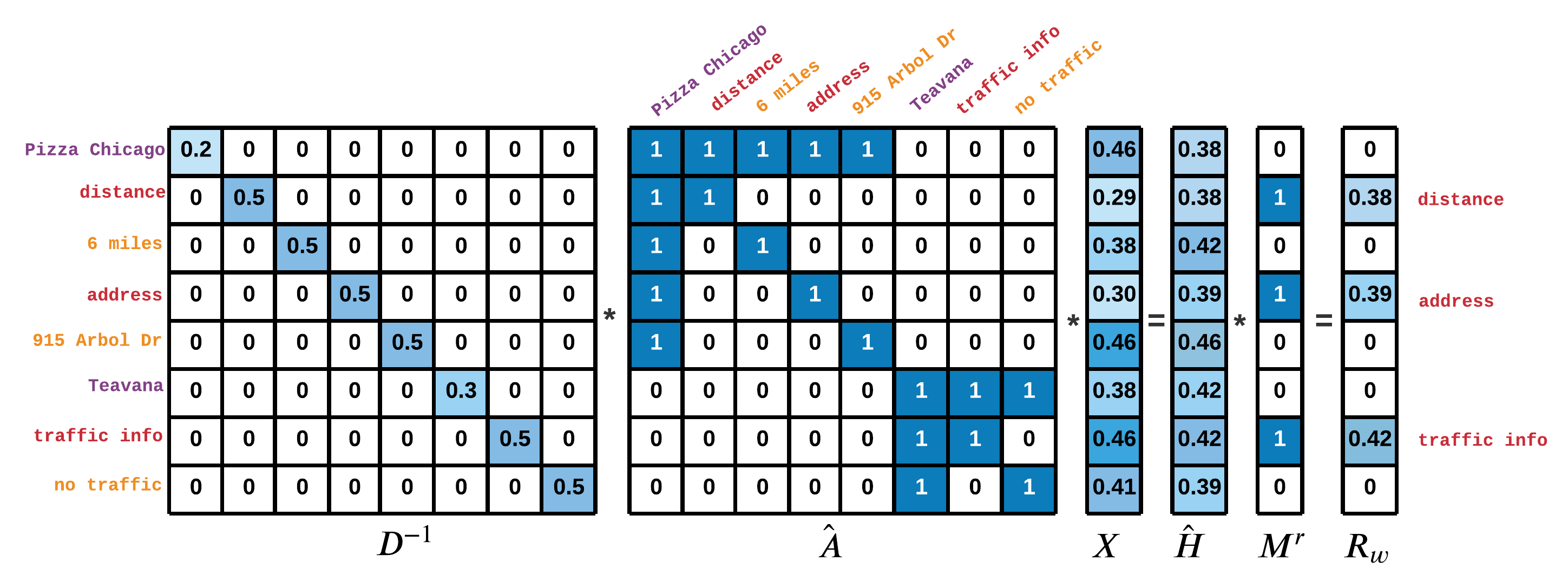}
    \caption{Relation weight computation.}
    \label{fig:graph_lap}
  \end{subfigure}

  \caption{For the graph in Figure (a) and the question "\textit{Find me the quickest route to the restaurant?}" the computation of the relation weight is shown in Figure (b), where $\hat{A} = A+I$.}

\end{figure*}
\textbf{Knowledge Specific Embedding.} To infuse structural information, DialoKG employs entity embedding, triple embedding and type embedding, besides the usual word token and positional embedding.  %(Figure~\ref{fig:enc}).
Such an embedding technique allows the system to encode the knowledge graph structure. To do this, knowledge graph triples are linearized into a sequence as input, as depicted in Figure~\ref{fig:enc}. To facilitate order invariance of the knowledge embedding, we shuffle the order of the graph triples in the input sequence during training. In the token embedding layer {\fontfamily{lmtt}\selectfont[S]}, {\fontfamily{lmtt}\selectfont[R]} and {\fontfamily{lmtt}\selectfont[O]} are special tokens to separate subject, relation and object of a triple from each other in the sequence. Entity and triple embedding layers embed entity and triple-level information of the word token. For instance, {\fontfamily{lmtt}\selectfont ENT1} in the entity embedding layer indicates that the corresponding words in the token embedding layer are related to the first subject, which is \textit{starbucks} in this case. Likewise, {\fontfamily{lmtt}\selectfont T1} and {\fontfamily{lmtt}\selectfont T2} in the triple embedding layer indicate that the corresponding words in the token embedding layer are related to the first and second triple, respectively.
Finally, the type embedding indicates that the corresponding tokens are from the knowledge graph as opposed to the dialogue history.

\textbf{Dialogue Specific Embedding.}
The dialogue specific part of the input sequence is separated from the knowledge specific part by a {\fontfamily{lmtt}\selectfont[SEP]} token in the token embedding layer. %Dialogue sequence  (Dialogue History + Question) is separated by a  in the token embedding layer. 
Furthermore, the user utterance/question ($\mathcal{Q}$) of the current turn is separated by a {\fontfamily{lmtt}\selectfont[Q]} token from the dialogue history. The type embedding layer %in the dialogue encoder 
stores information about whether the corresponding utterance is from the user or system.
%encodes information about the type of the speakers. % the model about which word token belongs to user question and system response, in the dialogue. 
%With the \textbf{USR} and \textbf{SYS} tokens, which corresponds to user and system utterances in the Type embedding layer, 
This way, the decoder can use information about typical dialogue turn patterns.
%that the speakers follow. 

The positional embedding in both knowledge and dialogue embeddings encodes the position of each word token in the sequence. Finally, embeddings from all five layers are summed up as depicted in Figure~\ref{fig:enc}. \textit{Layer Normalization}~\cite{layernorm} is then applied to obtain the final embedding representation of the complete input sequence. It normalizes the embedding representation of layers, which restricts the weights of the learning network from exploding.
%a follows:
%\begin{equation}
%    w^{emb} = LayerNorm(w)
%\end{equation}

%that includes knowledge graph and the dialogue context.
We argue that the proposed design pattern of forming a single sequence and specifying each item in the input sequence further with additional embedding layers can improve the system's understanding of the task and domain.

\subsection{Knowledge Attention Mask Construction}
\label{subsec:graphweight}
%A knowledge graph weighted-attention mask is constructed to inform the decoder about the relevant KG triples for answering the user question of the current dialogue turn.
To notify the decoder about the relevant KG triples for answering the current user question, a knowledge graph weighted-attention mask is constructed. Prior to the construction of the knowledge attention mask, a weighted-knowledge graph, $\mathcal{G}_{w}$ is first computed by a \textit{Graph Weight Computer} module, %(Figure~\ref{fig:graphwg})
where the entity and relation weights are computed independently. We discuss the components of the \textit{Graph Weight Computer} module below.

\paragraph{Entity Weight Estimator.}
A pre-trained language model RoBERTa~\cite{liu2019roberta}, is used to compute the entity weights, similar to~\cite{yasunaga-etal-2021-qa}. Each entity $E_{i}$ $\in$ $\mathcal{E}$ of graph $\mathcal{G}$ is concatenated with the user utterance $\mathcal{Q}$ to obtain the probability score from the language model.
\begin{equation}
    E_{iw} = LM_{head}(LM_{enc}([\mathcal{Q};E_{i}]))
    \label{eq:entw}
\end{equation}
In Equation~\ref{eq:entw}, ${LM}_{head} \circ LM_{enc}$ represents the probability of the entity $E_{i}$ computed by the language model.
We consider $E_{iw}$ as the weight of the entity $E_{i}$, which represents the relevance of the entity for the given user utterance $\mathcal{Q}$.
\paragraph{Relation Weight Estimator.}
\label{subsec:relweight}
%We obtain the relation weights using the concept of Graph Convolution following the work of. However, unlike the previous works where a trainable weight and a graph signal is used, use compute the relation weights in an unsupervised manner as follows:

We follow~\cite{kipf2017semi,vashishth19a} and leverage the concept of GCN to obtain the relation weight. In contrast to the previous works, our proposed relation weight estimator transforms the input graph into an undirected graph, where the relations are considered as nodes of a graph. This transformation technique allows the relation estimator to obtain a score for each relation. The graph transformation is demonstrated in Figure~\ref{fig:lapgraphmini}. The relation weight is computed as follows:

%by first computing the relation relevance vector, $R_{r}$ as follows:

\begin{equation}
\begin{split}
R_{w} &=  \Tilde{H}M^{r} ,
\\
\Tilde{H} &= D^{-1} (A+I) X\\
%X_{ij} &= \frac{KG_{w}^{emb}\cdot Q^{emb}}{\lVert{KG_{w}^{emb}}\rVert \lVert Q^{emb}\rVert}
\end{split}
\label{eq:graphlap}
\end{equation}
Here, $D^{-1} (A+I)$ computes the row-normalized adjacency matrix, where $D$ and $A$ are respectively the degree matrix and adjacency matrix of the graph $\mathcal{G}$ as depicted in Figure~\ref{fig:graph_lap} and $I$ is the identity matrix. Let $d_{g}$ = $|\mathcal{E}|$ + $|\mathcal{R}|$ be the total number of entities and relations in the graph $\mathcal{G}$, then $D,A,I\in \mathbb{R}^{d_{g}\times d_{g}}$.
A feature vector $X\in \mathbb{R}^{d_{g}\times 1}$ is obtained by computing the cosine similarity between the embedding of knowledge graph entries (entities and relations) and the embedding of question. Furthermore, a relation mask $M^{r}\in \mathbb{R}^{d_{g}\times 1}$ is constructed by setting a value of $1$ and $0$ to the positions that correspond to relations and entities, respectively, to attend to the values that correspond to the relations only.
%the relation position to $1$ and entity positions to $0$ (see Figure~\ref{fig:graph_lap}).
Finally, values that correspond to the entities in $\hat{H}$ are masked out by multiplying with $M^{r}$ to obtain final relation weights $R_{w}\in\mathbb{R}^{d_{g}\times d_{g}}$. 
\begin{table*}[!th]
\begin{adjustbox}{width=\textwidth,center}
\begin{tabular}{lccccc}
\toprule
\textbf{Dataset} & \multicolumn{1}{c|}{\textbf{\#Dialogues}} & \multicolumn{1}{c|}{\textbf{\#Utterances}} & \multicolumn{1}{c|}{\textbf{Avg. Length of Utt.}} & \multicolumn{1}{c|}{\textbf{\#Utt. with Entities}} & \multicolumn{1}{c}{\textbf{Avg. \#Entities per Utt.}} \\ \midrule
SMD~\cite{eric-etal-2017-key} & 3,031 & 15,928 & 9.22 & 4430 & 2.96 \\
CamRest~\cite{wen-etal-2017-network} & 676 & 2,744 & 11.72 & 2366 & 2.43 \\
MWOZ~\cite{budzianowski-etal-2018-multiwoz} & 2,877 & 19,870 & 16.68 & 6241 & 2.06\\
\bottomrule
\end{tabular}
\end{adjustbox}
\caption{Dataset statistics.}
\label{tab:datastat}
\end{table*}
The computed weighted-graph assists the model to focus on the task by constructing a knowledge attention mask. We use the normalized score of $R_{w}$ and $E_{w}$ for constructing the knowledge attention mask. To filter-out irrelevant knowledge triples, we obtain the top-$k$ entities and relations from the weighted graph as denoted as $\hat{\mathcal{E}}$ and $\hat{\mathcal{R}}$ respectively. Here, $k$ is a hyper-parameter which we chose from a range of [0, max(|$\mathcal{E}$|, |$\mathcal{R}$|)], based on the validation score. Finally, based on the selected $\hat{\mathcal{E}}$ and $\hat{\mathcal{R}}$, the knowledge attention mask is constructed as follows:
\[
    M_{i,j}^{kg}= 
\begin{dcases}
    0,& \text{if } ((s_{i}\vee o_{i})\in \hat{\mathcal{E}})\wedge(r_{i}\in \hat{\mathcal{R}})\\
    -\infty,              & \text{otherwise}
\end{dcases}
\]
%Let $M^{dg}$ be the mask for the dialogue related input tokens, then we pad $M^{kg}$ to match sequence length and construct the final attention mask as $M$=$[M^{kg};M^{dg}]$. Here, 

Here $r_{i}$, $s_{i}$ and $o_{i}$ corresponds to the relation, subject and object entity of triple $\mathcal{T}_{i}$. Any position that corresponds to the value of $-\infty$ results in $0$ after computing the \textit{softmax} during the attention computation (discussed in the next sub-section)% in Equation~\ref{eq:attn}
. The final mask $M\in \mathbb{R}^{n\times n}$ is obtained by appending the mask for the dialogue related sequence with the knowledge attention mask where $n$ is sequence length. Padding is added to adjust the dimension of the metrics.
%By appending and padding the knowledge mask $M^{kg}$ with the usual mask for the dialogue sequence, we obtain the final mask $M\in \mathbb{R}^{n\times n}$ where $n$ is sequence length. Here, $M$ is the attention mask that mask out the next tokens during the decoding step.

\subsection{Decoder}
A Transformer~\cite{vaswani2017attention} based GPT-2~\cite{radford2019language} model is used for generating the response. The attention, computed in each of GPT-2's heads is formalized as follows:
\begin{equation}
\small
\begin{split}
Attn(Q,K,V) &= softmax{(\frac{1}{\sqrt{d_{k}}}(QK^{T})+M)}V, \\
H_{i} &= Attn(QW^{Q}_{i},KW^{K}_{i},VW^{V}_{i})
%Y &= MLP([H_{1},...,H_{N_{h}}])
\end{split}
\label{eq:attn}
\end{equation}

where, $Attn(\cdot)$ computes the masked attention, $H_{i}$ is the $i$-th head, $d_{k}$=$d_m/h$. Here, $d_{m}$ is the dimension of the model where $h$ the number of heads. %$MLP(\cdot)$ is a feed forward network and Y is the output. Here, $Q$, $K$ and $V$ are computed form the input embedding obtained from \textsection{\ref{subsec:inpemb}}.
$Q$, $K$ and $V$ are query, key and value where $W^{Q}_{i}, W^{K}_{i}, W^{V}_{i}$
%$\in \mathbb{R}^{d_{h}\times d_{m}}$
are trainable parameters. The objective of the model is to minimize the negative log-likelihood $\mathcal{L}$ for next-token prediction. For a dialogue dataset $D=\{D_{1},D_{2},...,D_{j}\}$, we formally define $\mathcal{L}$ as follows: 
\begin{equation}
\small
\mathcal{L}(D) = -\sum_{j}^{|D|}\sum_{i}^{n} \text{log }p(s_{i}^{j}|s_{1}^{j},.,s_{i-1}^{j},\mathcal{H}^{j},\mathcal{Q}^{j},\mathcal{G}^{j}),
\end{equation}

where $n$ is the maximum response length and $\mathcal{H}^{j},\mathcal{Q}^{j},\mathcal{G}^{j}\in D_{j}$. Top-k sampling~\cite{fan2018hierarchical} decoding is used to generate the next word token at each time step, during the inference.

\section{Experimental Setup}

\subsection{Data}
We evaluate DialoKG on three publicly available knowledge-grounded and task-oriented dialogue datasets: Stanford Multi-Domain dataset (SMD)~\cite{eric-etal-2017-key}, CamRest~\cite{wen-etal-2017-network} and Multi-WOZ 2.1 (MWOZ)~\cite{budzianowski-etal-2018-multiwoz}. SMD consists of three domains: weather, navigation, and calendar. MWOZ contains five domains: train, hotel, restaurant, taxi and attraction. %Train/validation/test splits for the SMD, CamRest and MWOZ are 2,425/302/304, 406/135/135 and 1,839/117/141 dialogues respectively.
We use the splits provided with the datasets for train, validation, and test. Each dialogue is provided with a knowledge base. Table~\ref{tab:datastat} shows the statistics of the benchmark datasets.% More detail about the datasets are provided in the Appendix~\ref{app:data}.

\subsection{Hyper-parameter Settings}
Throughout this paper, we use the GPT-2~\cite{radford2019language} model with 117M parameters. AdamW~\cite{loshchilov2018decoupled} with $\epsilon$ = $1e$-$8$ and learning rate of $6.25e$-$5$ is employed as optimizer. GELU~\cite{hendrycks2016gaussian} is used as activation function. The best hyper-parameters for each dataset were found using grid search and based on the results on the validation set. We run all experiments on a distributed training setting with 10 GPUs, each with 12 GB of memory.
More implementation details can be found in Appendix~\ref{app:trainparams}.
%Details about the training hyper-parameters per dataset are reported in the Appendix~\ref{app:trainparams}. 
\begin{table*}[]
\begin{adjustbox}{width=\textwidth,center}
\begin{tabular}{lccccccccc}
\toprule
                & \multicolumn{3}{c}{\textbf{SMD}}                        & \multicolumn{3}{c}{\textbf{CamRest}}                    & \multicolumn{3}{c}{\textbf{MWOZ}}                    \\ \cline{2-10} 
\textbf{Model} & \textbf{BLEU} & \textbf{MoverScore} & \textbf{Ent. F1} & \textbf{BLEU} & \textbf{MoverScore} & \textbf{Ent. F1} & \textbf{BLEU} & \textbf{MoverScore} & \textbf{Ent. F1} \\ \midrule

        GLMP~\cite{wu2019global}        &          13.9      &           54.2          &      59.6              &       15.1        &              57.2       &     58.9               &      6.9         &           51.2          &       32.4             \\
        MLM~\cite{gangi-reddy-etal-2019-multi}        &     17.0          &        64.0             &       54.6             &    15.5           &           57.0          &         62.1           &         -      &       -              &         -           \\
        Ent. Const.~\cite{qin-etal-2019-entity}        &       13.9        &       53.8              &      53.7              &   18.5            &          65.9           &        58.6            &       -        &          -           &      -              \\
                GPT2+KE~\cite{madotto-etal-2020-learning}        &      17.4         &              \underline{66.4}       &          59.8          &           18.0   &           65.8          &               54.9     &        \textbf{15.0 }      &             \underline{60.9}        &           \underline{39.6}         \\
        TTOS~\cite{he-etal-2020-amalgamating}        &     17.4          &      59.8               &        55.4            &  20.5             &          67.0           &   61.5                 &       -        &          -           &    -                \\
        DF-Net~\cite{qin-etal-2020-dynamic}        &      14.4         &        56.3             &       62.7             &   -            &         -            &    -                &        9.4       &          54.2           &          35.1          \\
        EER~\cite{he2020task}        &           17.2    &            60.9         &     59.0               &     19.2          &       66.1              &        65.7            &       13.6        &         57.2            &  35.6   \\
                
        FG2Seq~\cite{9053667}                &        16.8       &           60.2          &             61.1       &         20.2      &            66.6         &                66.4    &      \underline{14.6}         &         58.4            &  36.5   \\
                        
        CDNet~\cite{raghu-etal-2021-constraint}                &    \underline{17.8}           &        61.1             &  \underline{62.9}                  &      \underline{21.8}         &          \underline{67.8}           &        \underline{68.6}            &          11.9     &        55.8             &  38.7   \\ \midrule
                \textbf{DialoKG}                &     \textbf{20.0}          &          \textbf{70.6}           &                \textbf{65.9}    &      \textbf{23.4}         &           \textbf{70.4}         &                    \textbf{75.6} &       12.6        &    \textbf{62.6}                 &  \textbf{43.5}   \\
                \bottomrule
\end{tabular}
\end{adjustbox}
\caption{Performance of DialoKG and baseline models on three benchmark datasets. Best scores in \textbf{bold} and second-best \underline{underlined}.}
\label{tab:allresults}

\end{table*}
\subsection{Evaluation Metrics}
\paragraph{Automatic Metrics.} Following the baseline models, we use BLEU~\cite{papineni2002bleu} and Entity F1 score~\cite{eric-etal-2017-key} as automatic evaluation metrics. The Entity F1 score represents the model's capability of generating knowledge grounded responses. It computes the F1 score between the set of entities present in the ground truth and system-generated responses. Several studies~\cite{novikova-etal-2017-need,liu-etal-2016-evaluate} on evaluation metrics suggest that word-overlap based metrics such as BLEU are insufficient for evaluating natural language generation (NLG) systems. Hence, we use MoverScore~\cite{zhao-etal-2019-moverscore} as addition metric to evaluate the semantic similarity between the system generated response and the ground truth. We compute both MoverScore and BLEU scores on the sentence level.

\paragraph{Human Evaluation.}
To assess the quality of the system-generated responses, we conduct a human evaluation based on the following criteria: 1) Naturalness: how human-like and fluent the generated responses are, and 2) Correctness: how correct the knowledge-grounded responses are. We asked three annotators (two from Computer Science (CS) and one from a non-CS background) who are not part of this research work to evaluate the quality of the system-generated responses. We randomly sampled 90 dialogues in total from the benchmark datasets and asked annotators to evaluate the system-generated responses given the ground truth response and the knowledge graph triples on a scale of [1,5] (higher is better). The inter-annotator agreement score (Cohen's kappa $\kappa$) of the annotated data is 0.82. The human evaluation process is explained in detail in Appendix~\ref{app:humeval}.

\subsection{Baselines}
We compare DialoKG with the following state-of-the-art methods: %Memory network oriented systems
\textbf{GLMP}~\cite{wu2019global}, \textbf{MLM}~\cite{gangi-reddy-etal-2019-multi}, \textbf{Ent. Const.}~\cite{qin-etal-2019-entity}, \textbf{DF-Net}~\cite{qin-etal-2020-dynamic}, \textbf{CDNet}~\cite{raghu-etal-2021-constraint}, %: A global memory encoder and a local memory decoder is used in GLMP to generate dialogues from external knowledge.
%\textbf{MLM}~\cite{gangi-reddy-etal-2019-multi} uses a multi-level memory architecture in a key-value setup for dialogue generation. 
%\textbf{Ent. Const.}~\cite{qin-etal-2019-entity}: A retrieval mechanism to fetch the relevant rows from the knowledge base and an attention based column selection method is used to filter appropriate knowledge during the dialogue generation. 
%Furthermore, we compare with 
\textbf{GPT2+KE}~\cite{madotto-etal-2020-learning}, 
%that uses a method to embed the knowledge directly into the model parameters to generate dialogues. A reinforcement based system, 
\textbf{TTOS}~\cite{he-etal-2020-amalgamating}
%. % uses a "Two-Teacher One-Student" method, where reinforcement learning is used to train the teacher networks in a goal-specific reward setting for the dialogue generation task.
%\textbf{DF-Net}~\cite{qin-etal-2020-dynamic} uses a dynamic fusion network that automatically exploits the relevance between different domains.
and \textbf{EER}~\cite{he2020task}.
%:  An enhanced entity representation technique is used to fetch context-sensitive and structure-aware entity representation during the dialogue generation process. \textbf{FG2Seq}~\cite{9053667} is a seq2seq model that encodes the structured knowledge and the latent semantic representation of the dialogue history.% \textbf{CDNet}~\cite{raghu-etal-2021-constraint}: A cosine similarity based knowledge distillation method is used in CDNet to filter relevant information for dialogue generation.
Most of these approaches adopt memory networks to generate knowledge grounded dialogues, whereas \textbf{GPT2+KE}~\cite{madotto-etal-2020-learning} directly embeds the knowledge base into the model's parameters and \textbf{TTOS}~\cite{he-etal-2020-amalgamating} proposed a reinforcement learning-based framework.

\section{Results and Analysis}
\label{sec:results}
\subsection{Quantitative Results}
We conduct both quantitative and qualitative analyses to assess system-generated responses. Table~\ref{tab:allresults} summarizes the performance of DialoKG with respect to the baseline models. %CamRest contains dialogues about restaurant reservation. 
It is evident that DialoKG outperforms the baseline models significantly in Entity F1 score on CamRest, which contains mostly knowledge-grounded dialogues about restaurant reservations. % (see Table~\ref{tab:datastat}). 
%We observe from Table~\ref{tab:datastat} that CamRest contains mostly knowledge-grounded dialogues. 
A high Entity F1 score of $75.6$ on CamRest shows DialoKG's ability to generate knowledge-grounded with high accuracy. Although DialoKG achieves an improved Entity F1 score on the MWOZ dataset, it has a lower BLEU score since MWOZ often contains lengthy responses. However, the high MoverScore across all datasets demonstrates that DialoKG can generate highly semantically similar responses. Domain-wise results are reported in Appendix~\ref{app:domrel} due to space limitation.

\begin{figure}[ht]
\centering
    \includegraphics[width=\columnwidth]{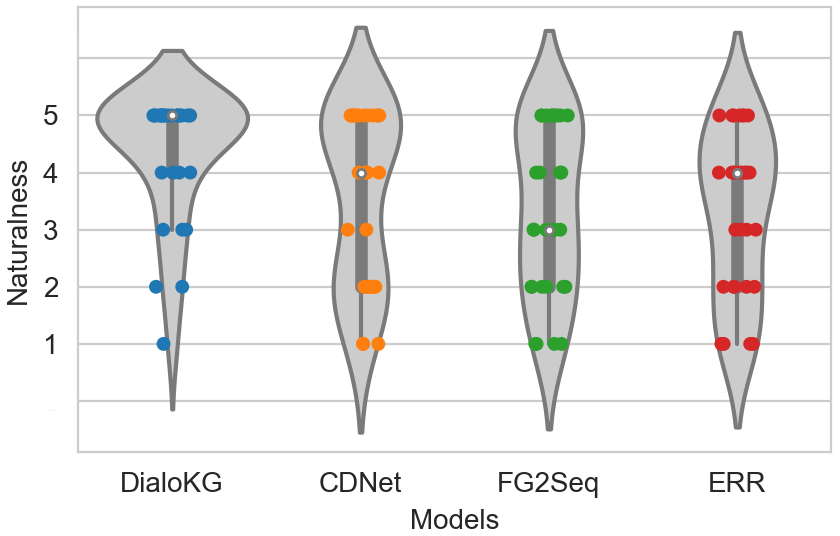}
    \caption{Distribution of human evaluation scores.}
    \label{fig:humscore}
\end{figure}

\begin{table}[]
\begin{adjustbox}{width=0.9\columnwidth,center}
\begin{tabular}{lcc}
\toprule
\textbf{Model}  & \textbf{Naturalness} & \textbf{Correctness} \\ \midrule
EER~\cite{he2020task}     &     3.27        &       3.61      \\
FG2Seq~\cite{9053667}    &     3.33        &       3.87      \\ 
CDNet~\cite{raghu-etal-2021-constraint}     &     3.53        &        3.94     \\ \midrule
\textbf{DialoKG} &      \textbf{4.33}       &  \textbf{4.01}   \\\bottomrule       
\end{tabular}
\end{adjustbox}
\caption{Human evaluation results.}
\label{tab:qualeval}

\end{table}

\subsection{Qualitative Results}
We obtain human evaluation scores (naturalness and correctness) for the closest three models. Results in Table~\ref{tab:qualeval} show that our proposed dialogue system can generate more human-like responses. An improved score is also achieved in terms of correctness, reflecting DialoKG's ability to generate highly accurate dialogues. Furthermore, Figure~\ref{fig:humscore} shows the distribution of human evaluation scores. The figure allows a better direct comparison of the individual score levels. 
%demonstrating how human-like (naturalness) the system-generated responses are. 
Details about the annotation process are reported in Appendix~\ref{app:humeval}.
%More experimental results are reported in Appendix~\ref{app:humeval}.

\begin{figure*}[ht]

\centering
    \includegraphics[width=\textwidth]{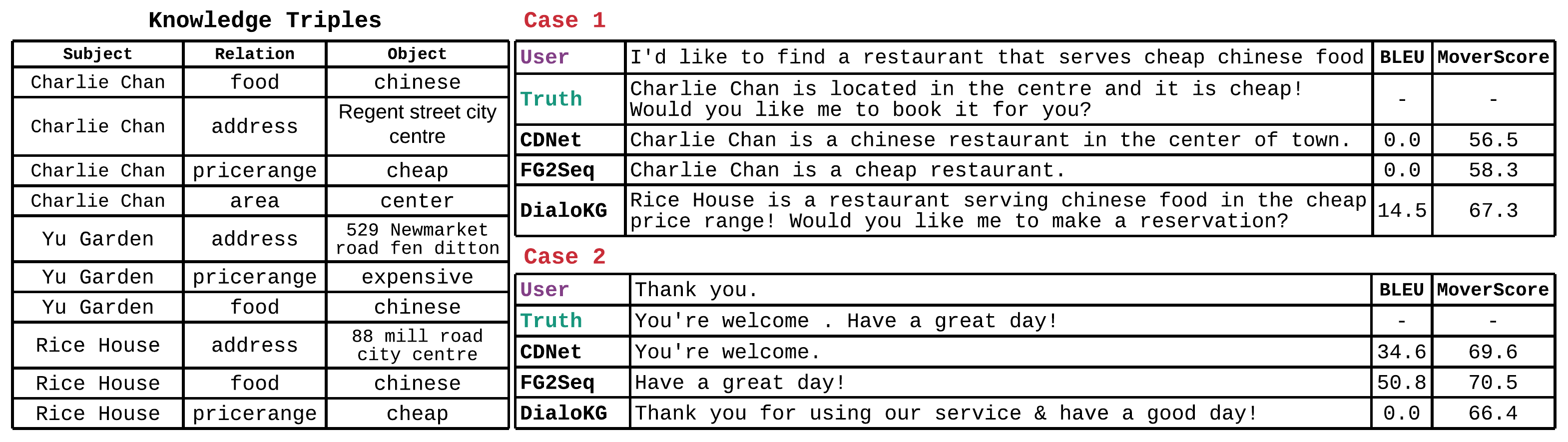}
    \caption{Case study: comparison between ground truth and system-generated responses.} 
    \label{fig:casestudy}
\end{figure*}
\begin{table}[]

\begin{adjustbox}{width=0.48\textwidth, center}
\begin{tabular}{lcccc}
\toprule
\textbf{Approach} & \textbf{BLEU} & $\Delta$ & \textbf{Ent. F1} & $\Delta$ \\ \midrule
\textbf{DialoKG} (seq2seq) & 14.5 & - & 59.4 & - \\
\quad + Entity embedding
 & 17.7 & 3.2\textcolor{teal!100!}{  \,$\uparrow$} & 63.0 &  3.6\textcolor{teal!100!}{  \,$\uparrow$}\\
 \quad + Triple embedding
 & 19.2 & 1.5\textcolor{teal!100!}{  \,$\uparrow$} & 67.8 &  4.8\textcolor{teal!100!}{  \,$\uparrow$}\\
 \quad + Type embedding
 & 20.1 & 0.9\textcolor{teal!100!}{  \,$\uparrow$} & 68.4 &  0.6\textcolor{teal!100!}{  \,$\uparrow$}\\
 \quad + Knowledge attention mask
 & 23.4 & 3.3\textcolor{teal!100!}{  \,$\uparrow$} & 75.6 &  7.2\textcolor{teal!100!}{  \,$\uparrow$}\\
 \bottomrule
\end{tabular}
\end{adjustbox}
\caption{Ablation study.}
\label{tab:ablation}
\end{table}

\subsection{Ablation Study}
We conducted an ablation study to investigate the contribution of major components of DialoKG. The results on CamRest in Table~\ref{tab:ablation} demonstrates that the \textit{ses2seq} approach achieves the lowest scores, which represents the DialoKG model without the embedding layers: entity embedding, triple embedding, and type embedding. %The results on CamRest in Table~\ref{tab:ablation} demonstrates that
Inclusion of the entity and triple embedding layers significantly improved model's performance in both BLEU and Entity F1 scores.
The type embedding further improved DialoKG's performance. The significant difference in results shows the effectiveness of the proposed embedding technique.
%each of the proposed components significantly improved the model's performance in both BLEU and Entity F1 score.
%Removing the knowledge attention mask and treating all the knowledge entries equally significantly decreases the Entity F1 score by 7.2 points. 
Finally, we observed a remarkable improvement in DialoKG's overall performance after the inclusion of knowledge attention mask. Question-aware weighted-graph computation used to construct knowledge attention mask, helped the model focus on the task at the inference time.%during the decoding process.
%Furthermore, without top-k entity and relation indicates selecting all the knowledge entries for constructing the knowledge mask adds irrelevant knowledge to the model.
%Furthermore, removing the entity, triple, and type embedding layers drops the results drastically. This significant difference in results shows the effectiveness of the proposed encoding technique.% The results in Table~\ref{tab:ablation} demonstrate the effectiveness of our proposed graph encoding and structure-aware decoding technique.

\subsection{Effectiveness of Knowledge Embedding}
The proposed graph embedding technique works best in combination with the knowledge attention mask. The graph embedding design allows DialoKG to handle disconnected graphs and triples. This makes DialoKG suitable for large-scale graphs, where a cosine-similarity based triple selection may be used to fit the graph triples inside the model's input capacity.
%The graph embedding technique is designed in such a way that it handles disconnected triples in a graph such as in Figure~\ref{fig:lapgraphmini}. 
The entity and triple embedding layers allow the model to preserve the structural information of a particular triple even though triples from different parts of the input sequence are selected based on the top-$k$ entities and relations to construct the knowledge attention mask. %the top-$k$ entities and relations are selected based on their weights for constructing the knowledge mask.
Overall, the graph embedding technique improves the Entity F1 score by 5.4, 9.0, and 3.7 points on SMD, CamRest, and MWOZ, respectively. This indicates the effectiveness of the proposed embedding techniques for capturing graph triples.
\begin{table}[]
\begin{adjustbox}{width=\columnwidth,center}
\begin{tabular}{ccccc}
\toprule
\textbf{Top-$k$ (entity)} & \textbf{Top-$k$ (relation)}  & \textbf{BLEU} & \textbf{MoverScore} & \textbf{Entity F1} \\ \hline
3              &  5              &       10.8        &         65.3            &          48.2          \\ \hline
3              & 7              &        11.0       &           65.4          &         48.9           \\ \hline
5              & 5              &        16.9       &           68.0          &           62.1         \\ \hline
5              & 7              &       17.4      &            68.1         &         62.5           \\ \hline
7              & 5              &       19.3     &             70.4        &         64.4           \\ \hline
7              & 7              &       \textbf{20.0}      &             \textbf{70.6}        &         \textbf{65.9}           \\ \hline
All            & All            &       15.9        &           67.2          &  59.0 \\\bottomrule                 
\end{tabular}
\end{adjustbox}
\caption{Effect of triple selection on the performance.}
\label{tab:topentrel}

\end{table}
\subsection{Impact of Knowledge Attention Mask}
To understand the effect of the knowledge-graph weighted attention mask, we experiment with the triple selection process described in DialoKG's approach. %\textsection{\ref{subsec:graphweight}}.
Table~\ref{tab:topentrel} shows the performance of DialoKG with selected top-$k$ entities and relations on the SMD dataset.  We observe that DialoKG achieves the best performance on SMD when the top 7 entities and relations are chosen to construct the knowledge mask. Consider the question "\textit{Do you have any local coffee shops?}" the ground truth is "\textit{There is Coupa, it s just 6 miles away but there is heavy traffic on our way}". The ground truth contains traffic information in addition to the distance and name of the coffee shop. Selecting a high number of entities and relations increases the chance of generating such additional information related to the subject of the question. However, choosing too many entities harms the model since it is more likely to add irrelevant noise (see Table~\ref{tab:topentrel}). For MWOZ, six entities and seven relations, and for CamRest, seven entities and five relations result in the best performance.%For MWOZ, six entities and seven relations are selected and for CamRest seven entities and five relations to achieve the best performance. 

\section{Case Study}
Figure~\ref{fig:casestudy} shows two cases from the MWOZ dataset given a subset of the knowledge graph. In Case 1, we observe that in answering the user question, DialoKG correctly picked \textit{Rice House} that serves \textit{cheap} and \textit{Chinese} food. However, in this case, multiple correct answers exist, e.g. \textit{Charlie Chan} also falls into the same category of restaurant. Despite generating the correct answer based on the given knowledge and the user question, DialoKG receives a low Entity F1 score since the generated response entity does not match the ground truth. In Case 2, where the baseline systems focus on imitating the ground truth, DialoKG generates a fluent and engaging response. Despite generating a meaningful and semantically similar sentence, it obtained a BLEU score of $0.0$ because of the low overlap with the ground truth response. However, a high MoverScore in both cases indicates DialoKG's ability to generate a semantically similar response. Overall, we observe that DialoKG can generate human-like, engaging, and informative responses in a multi-turn dialogue setting.

\subsection{Influence of Dialogue History}
\begin{figure}[ht]
\centering
    \includegraphics[width=\columnwidth]{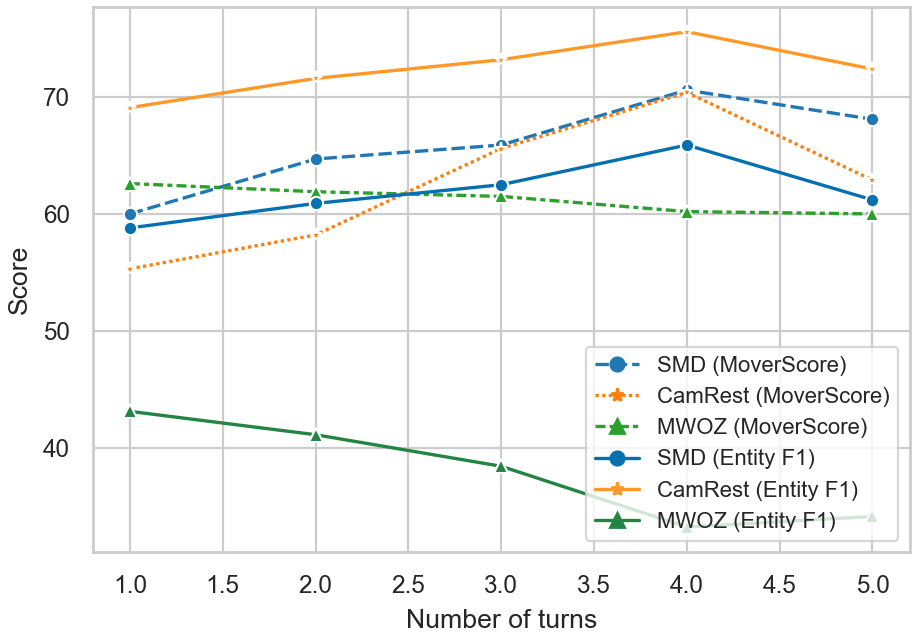}
    \caption{DialoKG's performance on benchmark datasets for different number of dialogue contexts.} 
    \label{fig:effhist}
\end{figure}

%Figure~\ref{fig:effhist} depicts the influence of dialogue history on the models' performance. 
Dialogue history is particularly crucial since it gives the model the context for generating the response. In some cases where the entity information is missing in the current user utterance, the dialogue context provides the model with enough information to perform the inference and generate the correct response. For instance, for the question, \textit{What is the food type they serve?}, the name of the restaurant is not given in the question, but the system can infer it from the dialogue history. However, from the experiments, we found that too much dialogue context may inject noisy and irrelevant information to answer the current question, in particular for knowledge-grounded responses in MWOZ. To quantify this, we selected different numbers of dialogue turns as history for the model's input depending on the characteristics of the dataset and visualised the result in Figure~\ref{fig:effhist}.

\section{Related Work}
\paragraph{Task-Oriented Dialogue Systems.}
Recent dialogue systems mainly leverage Memory Pointer Networks~\cite{memnet,mem2seq,wu2019global}, Copy mechanisms~\cite{gu-etal-2016-incorporating,lin-etal-2020-generating,chaudhuri2019using} and similarity-based knowledge distillation techniques~\cite{wen-etal-2018-sequence,raghu-etal-2021-constraint} for the knowledge selection and dialogue generation task. In this research direction, learning to generate template responses and fill in the slot is a common practice~\cite{wu2019global}. Dialogue history and knowledge entities are stored in shared memory, facilitating these systems to apply copy mechanisms over the memory space. A multi-level memory architecture is proposed by~\cite{gangi-reddy-etal-2019-multi} that handles the dialogue history and knowledge entries separately.

\paragraph{Knowledge-Structure Aware Dialogue Generation.}
Recently, several knowledge-grounded dialogue systems have attempted to capture structural knowledge to improve the performance of dialogue generation. A % structure-aware
sequence-to-sequence model is proposed by~\cite{liu2018table} that employs global and local attention to understand structural information. %which tries to capture the structural information of the knowledge entries by global and local attention.
Several studies found GCN to be effective for modeling graph-based data. Hence, they construct a graph from a document~\cite{moghe-etal-2020-incorporating} or the interaction between two speakers~\cite{wu2021csagn} for generating informative dialogues. %Then they train a model to generate dialogues by learning from the pattern of the constructed graph. 
In a different approach, an enhanced entity representation is proposed by~\cite{he2020task} by considering the entity information and the structural and relational information of the knowledge entries. In contrast to the previous works, our proposed approach represents the graph's structural information such as entities and triples through multiple embedding layers.

\paragraph{Language Model Based Dialogue Generation.}
Pre-training a language model with dialogue datasets~\cite{zhang2019dialogpt,bao-etal-2020-plato,gu-etal-2021-pral} and fine-tuning an already pre-trained model for various dialogue-related sub-tasks such as dialogue state tracking, action decision and response generation~\cite{hosseini2020simple,wu-etal-2020-tod,galetzka-etal-2021-space} has received much attention in recent years. Recently, \cite{madotto-etal-2020-learning} proposed a new method to embed the knowledge into the language model parameters. However, the authors noticed that the generated dialogues are sometimes noisy and requires high fine-tuning costs.  %However, the authors noticed that the generated dialogues are sometimes noisy and, in the case of dynamic knowledge, require high fine-tuning costs for updating the knowledge base. 
Despite the success of language model-based approaches, integrating structured knowledge into the dialogue generation process remains a challenging task. 
Unlike the previous approaches, we designed a structure-aware embedding method and exploit GPT-2 to generate dialogues.%both structure-aware graph encoding and decoding.

\section{Conclusion}
We have presented DialoKG, a novel knowledge-grounded task-oriented dialogue system improving the state-of-the-art across multiple benchmark datasets. DialoKG focuses on capturing the underlying semantics of the knowledge graph and pays attention to the relevant graph triples to understand the task and generate correct and human-like responses. The key contributions of DialoKG include 1) \textbf{Knowledge embedding technique}, that embeds the structural information of a knowledge graph effectively, and 2) \textbf{Knowledge graph-weighted attention masking}, which guides the masked language model to attend to the relevant knowledge entries for generating correct and informative responses. Finally, we showed DialoKG's ability to generate accurate, diverse, and human-like dialogues through quantitative and qualitative analysis. We performed an ablation study and studied the effect of dialogue history, knowledge embedding and knowledge attention masking.

\section*{Acknowledgements}
We acknowledge the support of the following projects: SPEAKER (BMWi FKZ 01MK20011A), JOSEPH (Fraunhofer Zukunftsstiftung), OpenGPT-X (BMWK FKZ 68GX21007A), the excellence clusters ML2R (BmBF FKZ 01 15 18038 A/B/C), ScaDS.AI (IS18026A-F) and TAILOR (EU GA 952215). The authors also acknowledge the financial support by the Federal Ministry for Economic Affairs and Energy of Germany in the project CoyPu (project number 01MK21007G).

\bibliography{anthology,custom}
\bibliographystyle{acl_natbib}

\appendix
\newpage

\section{Hyper-parameter Settings}
\label{app:trainparams}

We report the hyper-parameters used to train DialoKG in Table~\ref{tab:trainparamstab} for SMD, CamRest, and MWOZ. GPT-2 specific hyper-parameters are also reported in Table~\ref{tab:trainparamstab}. All the hyper-parameters are found after a grid search and evaluation on the validation set. We sample learning rate from \{6.25e-01, 6.25e-04, 6.25e-05\} and maximum history token and knowledge token from \{128, 256, 384, 512\}.
\begin{table}[h!]
\begin{adjustbox}{width=\columnwidth,center}
\begin{tabular}{lccc}
\toprule
 & \textbf{SMD} & \textbf{CamRest} & \textbf{MWOZ} \\\midrule
Learning rate &     6.25e-05         &             6.25e-04     &    6.25e-05           \\
Adam epsilon &     1e-08         &  1e-08      &    1e-08      \\
Batch size &        4      &          4        &         4      \\
Gradient accumulation steps &       4       &       4           &         4      \\
Max history turn &        4      &          4        &         1      \\
Maximum history token &        128      &              256   &         128      \\
Maximum knowledge token &        384      &          256      &         384      \\
Top relations &        7      &          7        &         6      \\
Top entities &        7      &       5           &   7   \\ 
Epochs &     40         &        25          &     30 \\
\bottomrule        
\end{tabular}
\end{adjustbox}
\caption{Training parameters.}
\label{tab:trainparamstab}
\end{table}

For both training and evaluation, we use a batch size of 4. Hyper-parameters used during the inference are reported in Table~\ref{tab:decparams}. We used 12 NVIDIA TitanX GPUs, each with 12GB of memory to train models. It took 30, 18 and 45 minutes to train on SMD, CamRest and MWOZ data.
\begin{table}[h!]
\begin{adjustbox}{width=0.9\columnwidth,center}
\begin{tabular}{lccc}
\toprule
 & \textbf{SMD} & \textbf{CamRest} & \textbf{MWOZ} \\\midrule
Temperature &      0.68        &         0.85         &          0.18     \\
Top-k &     6         &      8            &        10       \\
Top-p &     0.9         &      0.9            &     0.9          \\
Maximum response length &      100        &    80     &      120     \\
Top entities &     7         &    7     &    6  \\
Top relations &        7      &    5     &    7  \\
\bottomrule        
\end{tabular}
\end{adjustbox}
\caption{Decoding parameters.}
\label{tab:decparams}
\end{table}
\begin{table*}[!t]
\centering
\begin{adjustbox}{width=0.85\textwidth,center}
\begin{tabular}{l|ccc|ccc}
\toprule
\textbf{Models} & \textbf{BLEU} & \textbf{MoverScore} & \textbf{Entity F1} & \textbf{Schedule} & \textbf{Navigate} & \textbf{Weather} \\\midrule
GLMP~\cite{wu2019global} & 13.9 & 54.2 & 59.6 & 72.5 & 54.6 & 56.5 \\
MLM~\cite{gangi-reddy-etal-2019-multi} & 17.0 & 64.0 & 54.6 & 66.7 & 46.9 & 56.0 \\
Ent. Const.~\cite{qin-etal-2019-entity} & 13.9 & 53.8 & 53.7 & 55.6 & 54.5 & 52.2 \\
GPT2+KE~\cite{madotto-etal-2020-learning} & 17.4 & 66.4 & 59.8 & 72.6 & 53.5 & 57.7 \\
TTOS~\cite{he-etal-2020-amalgamating} & 17.4 & 59.8 & 55.4 & 63.5 & 45.9 & 64.1 \\
DF-Net~\cite{qin-etal-2020-dynamic} & 14.4 & 56.3 & 62.7 & 73.1 & 57.9 & 57.6 \\
EER~\cite{he2020task} & 17.2 & 60.9 & 59.0 & 71.8 & 52.5 & 57.8 \\
FG2Seq~\cite{9053667} & 16.8 & 60.2 & 61.1 & 73.3 & 56.1 & 57.4 \\
CDNet~\cite{raghu-etal-2021-constraint} & 17.8 & 61.1 & 62.9 & 75.4 & 56.7 & 61.3 \\ \hline
\textbf{DialoKG (Ours)} & \textbf{20.0} & \textbf{70.6} & \textbf{65.9} & \textbf{77.9} & \textbf{58.4} & \textbf{72.7}\\
\bottomrule
\end{tabular}
\end{adjustbox}
\caption{Domain-wise results on SMD dataset.}
\label{tab:domsmd}
\end{table*}

\section{Results}
\label{app:domrel}
We report the domain-wise results for SMD and MWOZ in Table~\ref{tab:domsmd} and Table~\ref{tab:domwoz} respectively. Baseline model's results are reported from~\cite{raghu-etal-2021-constraint} and~\cite{madotto-etal-2020-learning}. The MWOZ dialogue dataset contains conversations on the following domains as reported in the baseline works: attraction, restaurant, and hotel. The domain-wise results demonstrate that DialoKG achieves improved performance in almost all domains in a multi-domain setup. This demonstrates DialoKG's capacity to handle a dynamic knowledge base.

\begin{table*}[!t]
\centering
\begin{adjustbox}{width=0.85\textwidth,center}
\begin{tabular}{l|ccc|ccc}
\toprule
\textbf{Models} & \textbf{BLEU} & \textbf{MoverScore} & \textbf{Entity F1} & \textbf{Attraction} & \textbf{Restaurant} & \textbf{Hotel} \\\midrule
GLMP~\cite{wu2019global} & 6.9 & 51.2 & 32.4 & 24.4 & 38.4 &  28.1\\
MLM~\cite{gangi-reddy-etal-2019-multi} & - & - & - & - & - & - \\
Ent. Const.~\cite{qin-etal-2019-entity} & - & - & - & - & - & - \\
GPT2+KE~\cite{madotto-etal-2020-learning} & \textbf{15.0} & 60.9 & 39.6 & \textbf{43.3} & 37.1 & 33.4 \\
TTOS~\cite{he-etal-2020-amalgamating} & - & - & - & - & - & - \\
DF-Net~\cite{qin-etal-2020-dynamic} & 9.4 & 54.2 & 35.1 & 28.1 & 40.9 & 30.6 \\
EER~\cite{he2020task} & 13.6 & 57.2 & 35.6 & 43.0 & 34.3 & 35.7 \\
FG2Seq~\cite{9053667} & 14.6 & 58.4 & 36.5 & 37.2 & 38.9 & 34.4 \\
CDNet~\cite{raghu-etal-2021-constraint} & 11.9 & 55.8 & 38.7 & 38.9 & 41.7 & 36.3 \\ \hline
\textbf{DialoKG (Ours)} & 12.6 & \textbf{62.6} & \textbf{43.5} & 39.8 & \textbf{46.7} & \textbf{37.9}\\
\bottomrule
\end{tabular}
\end{adjustbox}
\caption{Domain-wise results on MWOZ dataset.}
\label{tab:domwoz}
\end{table*}
%\newpage\nwepage
\section{Knowledge Triples to Sequence Transformation}
\label{app:graphencdetail}
\begin{figure*}[ht]
    \includegraphics[width=\textwidth]{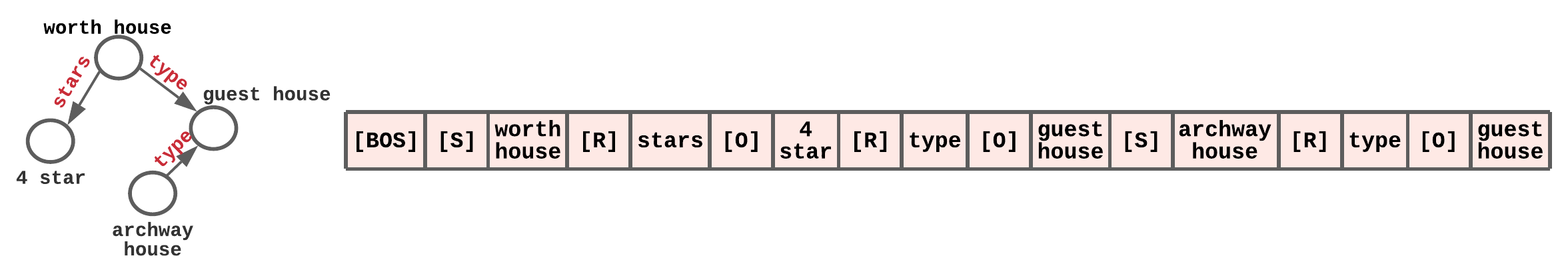}
    \caption{Illustration of graph to sequence transformation.}
    \label{fig:graphencill}
\end{figure*}

Figure~\ref{fig:graphencill} depicts how we linearize a graph into a sequence. The sequence begins with a [BOS] token, followed by the token [S] and a subject (\textit{worth house}). The token [S] indicates that the following word in the sequence is a subject (in this case \textit{worth house}). Then we append all the triples that are connected to the subject \textit{worth house} where the relation and object is separated by the token [R] and [O], respectively. Similarly, the second subject is appended to the sequence separated by a preceding [S] token.

%\section{Data}
%\label{app:data}

\section{Human Evaluation}
\begin{figure*}[h!]
    \includegraphics[width=\textwidth]{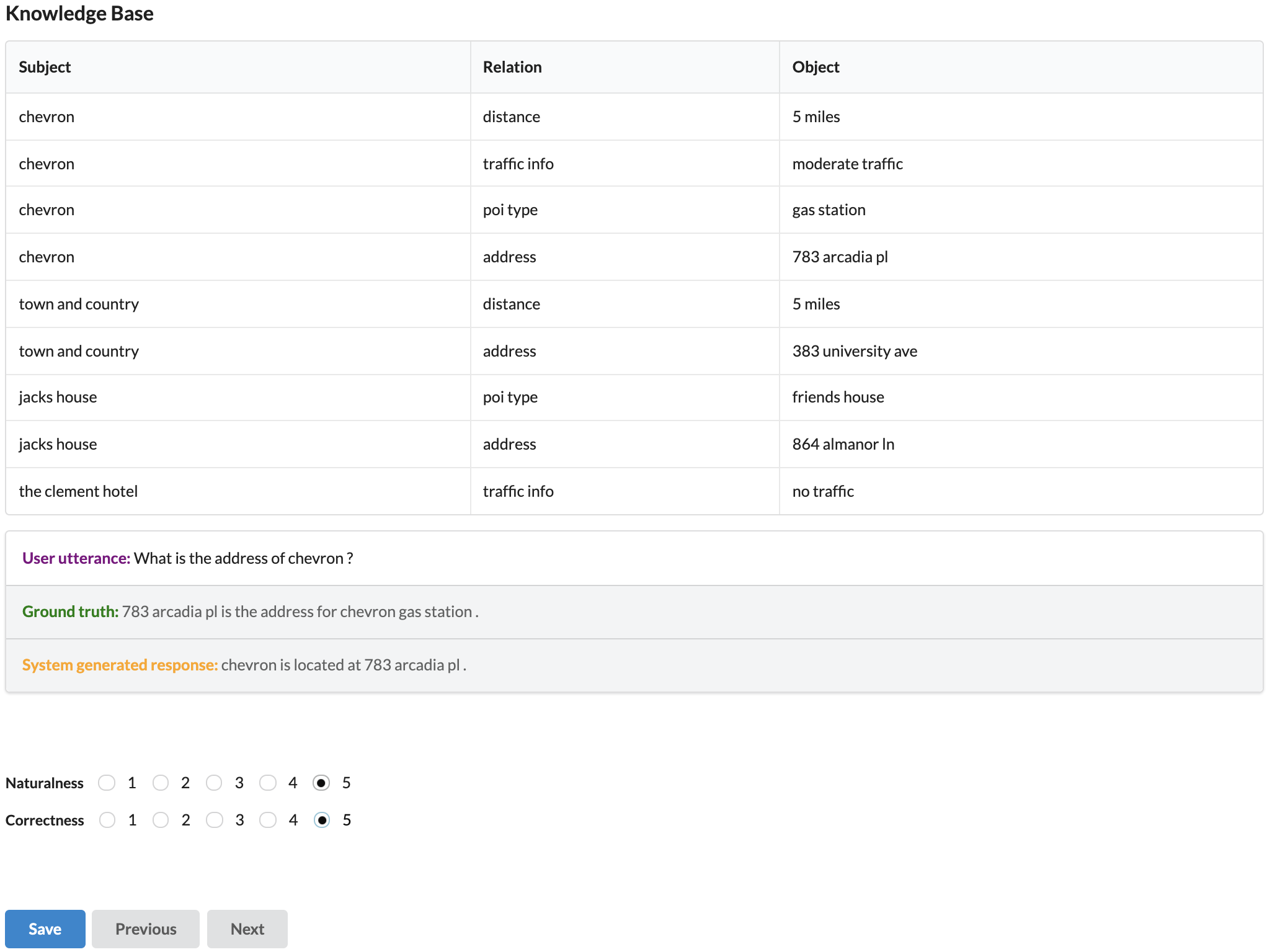}
    \caption{The interface of the annotation tool to obtained the human annotation scores.}
    \label{fig:humanntool}
\end{figure*}

Figure~\ref{fig:humanntool} shows the interface of the annotation tool used to obtain human annotation scores. The interface displays a set of knowledge triples, a user utterance, the ground truth response, and a system-generated response for each point. Given the information displayed on the annotation tool, we asked the annotators to rate the system-generated responses against the ground-truth on a scale of [1,5] (higher is better). We explained the participants about the purpose of this research. The first two participants are male (over 30 years old), and the third participant is female (more than 35 years old), both with several years of experience in the domain.
\label{app:humeval}

\section{Example System Outputs}
\begin{figure}[h!]
    \includegraphics[width=\columnwidth]{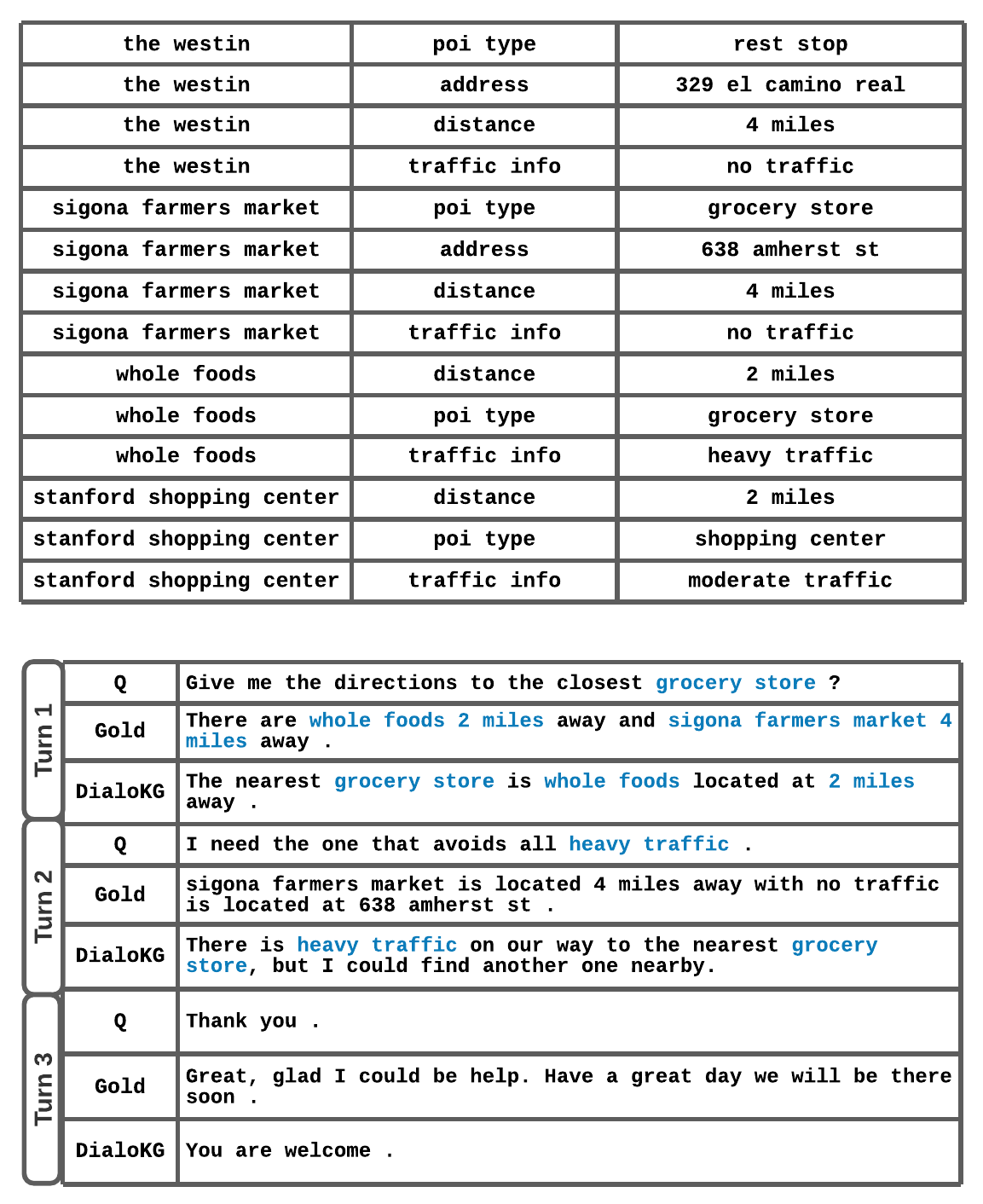}
    \caption{An example dialogue performed by DialoKG on SMD dataset.}
    \label{fig:samplesmd}
\end{figure}

\begin{figure}[h!]
    \includegraphics[width=\columnwidth]{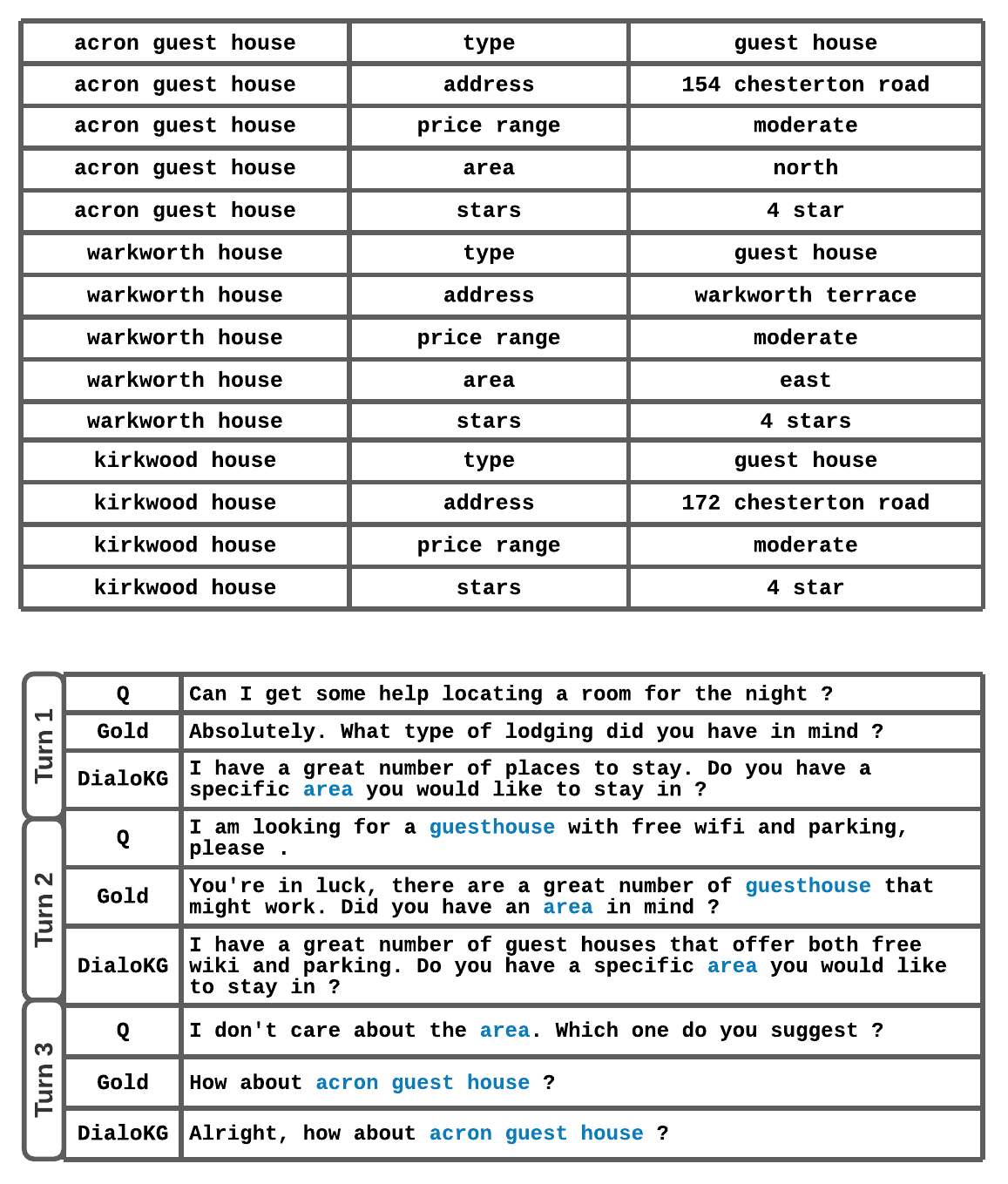}
    \caption{An example dialogue performed by DialoKG on MWOZ 2.1 dataset.}
    \label{fig:samplewoz}
\end{figure}

We show conversations performed by DialoKG on SMD and MWOZ 2.1 dataset in Figure~\ref{fig:samplesmd} and Figure~\ref{fig:samplewoz}, respectively. The example conversations demonstrate that DialoKG can perform accurate and engaging conversations.
\label{app:exmout}

%\section{User Case: MWOZ}
%\label{app:multi}

\end{document}